\documentclass{article} % For LaTeX2e
\usepackage{iclr2026_conference_arxiv,times}

\usepackage{amsmath,amsfonts,bm}

\def\eqref#1{equation~\ref{#1}}
\def\1{\bm{1}}

\DeclareMathAlphabet{\mathsfit}{\encodingdefault}{\sfdefault}{m}{sl}
\SetMathAlphabet{\mathsfit}{bold}{\encodingdefault}{\sfdefault}{bx}{n}

\usepackage{url}
\usepackage{graphicx}
\usepackage{booktabs}
\usepackage{graphicx}
\usepackage{enumitem}
\usepackage{pifont}
\usepackage[
pagebackref=true,
breaklinks,
colorlinks,
citecolor=blue,
linkcolor=blue,
urlcolor=blue,
hypertexnames=false
]{hyperref}
\usepackage{makecell}
\usepackage{cleveref}
\crefname{figure}{Fig.}{Figs.}
\Crefname{figure}{Figure}{Figures}
\crefname{section}{Sec.}{Secs.}
\Crefname{section}{Section}{Sections}
\Crefname{table}{Table}{Tables}
\crefname{table}{Tab.}{Tabs.}

\newcommand{\ql}[1]{\textcolor{black}{#1}}
\newcommand{\dz}[1]{\textcolor{black}{#1}}

\title{DiT360: High-Fidelity Panoramic Image Generation via Hybrid Training}

\author{%
\vspace{0.5em}%
Haoran Feng\textsuperscript{1,2}\footnotemark[1]~\quad
Dizhe Zhang\textsuperscript{1}\footnotemark[1]\footnotemark[2]~\quad
Xiangtai Li\textsuperscript{3}~\quad
Bo Du\textsuperscript{4}~\quad
Lu Qi\textsuperscript{1,4}\ding{41}\\
\vspace{0.5em}\small
\textsuperscript{1} Insta360 Research\quad
\textsuperscript{2} Tsinghua University\quad
\textsuperscript{3} Nanyang Technological University\quad
\textsuperscript{4} Wuhan University\\
}

\begin{document}

\maketitle

\let\thefootnote\relax\footnotetext{
$^*$ Equal contribution \hspace{5pt}
$^\dagger$ Project lead \hspace{5pt} 
\ding{41} Corresponding author
}

\begin{figure}[h]
    \centering
    \includegraphics[width=\linewidth]{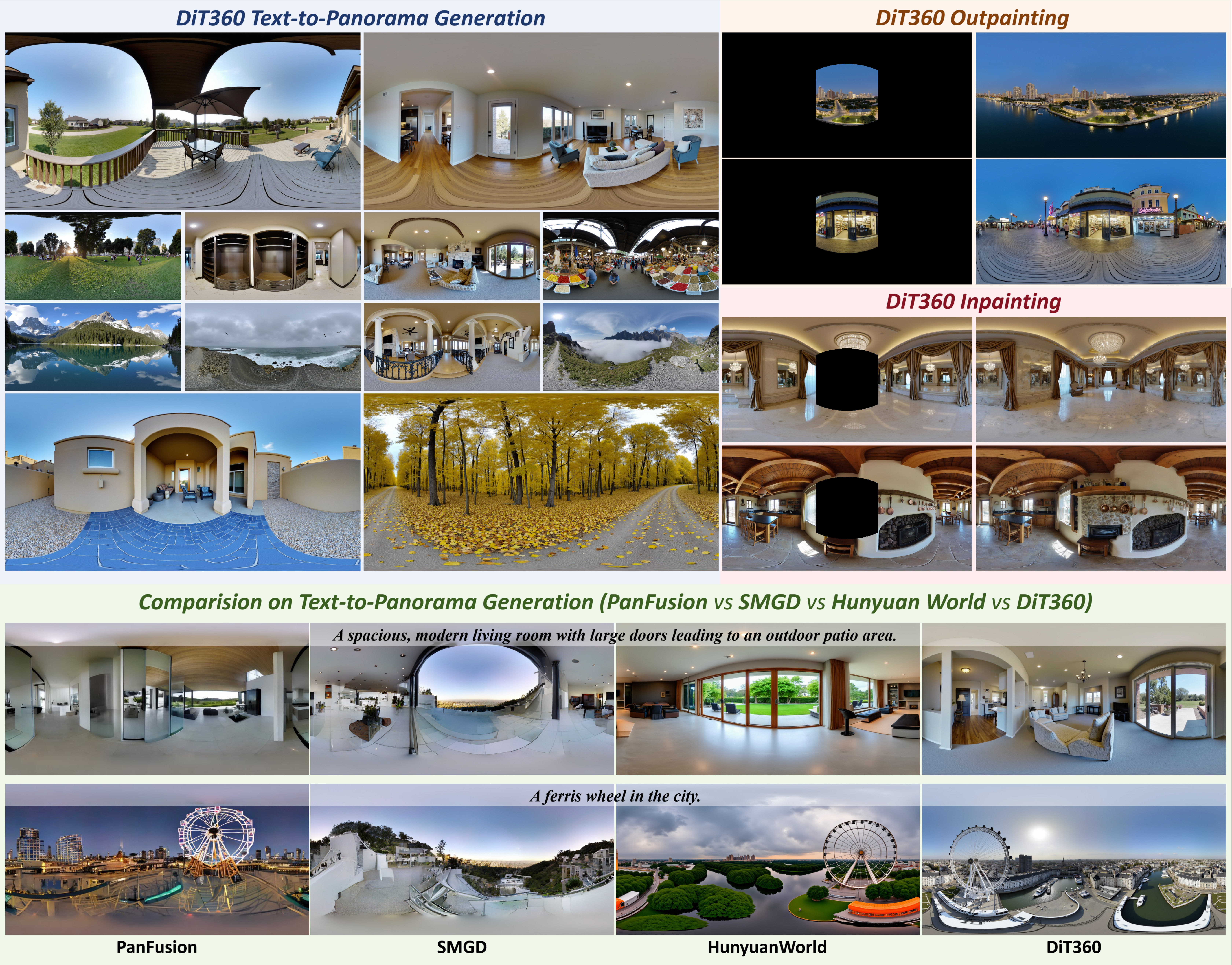}
    \caption{Visualization of \textit{DiT360}'s results. The shown examples include text-to-panorama generation, inpainting, and outpainting, together with comparisons against existing methods.}
    \label{fig:teaser}
\end{figure}

\begin{abstract}
In this work, we propose \textit{DiT360}, a DiT-based framework that performs hybrid training on perspective and panoramic data for panoramic image generation.
For the issues of maintaining geometric fidelity and photorealism in generation quality, we attribute the main reason to the lack of large-scale, high-quality, real-world panoramic data, where such a data-centric view differs from prior methods that focus on model design.
Basically, \textit{DiT360} has several key modules for inter-domain transformation and intra-domain augmentation, applied at both the pre-VAE image level and the post-VAE token level.
At the image level, we incorporate cross-domain knowledge through perspective image guidance and panoramic refinement, which enhance perceptual quality while regularizing diversity and photorealism.
At the token level, hybrid supervision is applied across multiple modules, which include circular padding for boundary continuity, yaw loss for rotational robustness, and cube loss for distortion awareness.
Extensive experiments on text-to-panorama, inpainting, and outpainting tasks demonstrate that our method achieves better boundary consistency and image fidelity across eleven quantitative metrics.
Our code is available at \href{https://github.com/Insta360-Research-Team/DiT360}{https://github.com/Insta360-Research-Team/DiT360}.

\end{abstract}
\section{Introduction}
\label{sec:intro}

% \lxt{Overall comments:
%     After I look at the entire draft. There are two significant issues.
%     The most crucial issue is unclear motivation for technical contributions. I think we need to separate the training methods, proposed loss, and padding methods. In addition, in the intro part, there should be clearer words to support these motivations.

%     The second issue is, please make the writing logic consistent between the intro, method, and conclusion.
% }

With the growing demand for spatial intelligence~\citep{yang2025thinking,chen2024spatialvlm,wu2025spatial}, panoramic image generation~\citep{lin2025one} has become critical considering its ability to capture the full 360° field of view.
Unlike conventional image generation on perspective views~\citep{sd3,sdxl,VAR,flux}, panoramic image generation remains challenging due to its unique characteristics such as severe distortions in polar regions, which in turn hinders its wider deployment in applications such as AR/VR~\citep{wang2025holigs,shi2025dreamrelation} and autonomous driving~\citep{yue2025roburcdet,qi2019amodal}.

To address this issue, existing methods usually focus on specific model design based on equirectangular projection (ERP)~\citep{ye2024diffpano,huang2025dreamcube,kalischek2025cubediff,panfusion,worldgen,smgd,hunyuanworld,ni2025makes,smgd}, with or without the assistance of cubemaps (CP)~\citep{bar2023multidiffusion,li2023panogen,shi2023mvdream,mvdiffusion,park2025spherediff,yang2025omni}, where both ERP and CP are common panoramic representations.

Despite the success achieved, these works still struggle with perceptual realism and geometric fidelity due to the scarcity of high-quality real-world panoramic data and the over-reliance on simulated one.

A heuristic solution is to exploit 360° data from media platforms such as YouTube, but its direct use for training is impractical since panoramic data requires domain-specific curation, including horizon correction and aesthetic filtering, which remain largely unexplored. Thus, one question raised: \textit{how can models be endowed with real-world knowledge when only limited panoramic data is available?}

By this motivation, we propose DiT360, a DiT-based framework~\citep{dit}, which adopts a hybrid training strategy that combines limited synthetic panoramic data with well-curated, high-quality perspective images to enhance photorealism and geometric fidelity simultaneously.
To fully realize the \ql{merit} of this hybrid paradigm, it is essential to leverage knowledge from the two domains at different representation levels. Accordingly, the \textit{DiT360} incorporates several key modules for inter-domain transformation and intra-domain augmentation, applied at both the pre-VAE image level and the post-VAE token level.
At the image level, the focus is on regularization across different domains, where existing panoramic data is regularized through masking and inpainting to remove spatial-variant artifacts in polar regions, while perspective data is regularized into the panoramic space through projection-aware methods to provide photorealistic guidance.
At the token level, the focus is on geometry-aware supervision in the latent space. Circular padding aims to address the boundary continuity problem of ERP images, where the left/right edge correspond to inherently periodic 0°/360° longitude. In addition, global rotational consistency is enforced through rotation-consistent yaw loss, while distortion-aware cube loss provides complementary supervision beyond ERP that guides the model toward consistent and high-fidelity panoramic representations.

The extensive experiments demonstrate that
\textit{DiT360} with hybrid training can perform better than existing text-to-panorama methods in boundary consistency and image fidelity, as evidenced by both quantitative metrics and qualitative visualizations.
For example, DiT360 achieves state-of-the-art performance on the Matterport3D validation set, surpassing prior methods across nine metrics like FID, Inception Score, and BRISQUE.
Beyond text-to-panorama generation, \textit{DiT360} naturally supports inpainting and outpainting tasks without additional finetuning enabled by its built-in masking and inpainting strategy.
Furthermore, our method can
produce high-resolution and photo-realistic panoramic images benefited by high-quality perspective data.
Our main contributions are summarized as follows:
\begin{itemize}[noitemsep, leftmargin=*]
    \item We present \textit{DiT360}, a DiT-based framework with hybrid training that leverages both perspective and panoramic data to preserve photorealism and geometric fidelity. Unlike prior approaches that primarily focus on model design, \textit{DiT360} emphasizes the effective utilization of multi-domain data to achieve superior generation quality.
    \item The proposed hybrid paradigm is realized through multi-level mechanisms, where image-level regularization refines existing panoramas and leverages perspective data to enhance diversity and photorealism, while token-level supervision in the latent space enforces geometric consistency through rotation- and distortion-aware constraints.
    \item Extensive quantitative and qualitative experiments on three tasks including text to image, inpainting and outpainting demonstrate that \textit{DiT360} outperforms existing methods in boundary consistency, image fidelity, and overall perceptual quality. 
    The user study conducted further confirms that our method aligns better with human preferences. 
\end{itemize}

\section{Related Work}
\label{sec:related_work}

\textbf{Text-to-Image Diffusion Models.}
Diffusion models have replaced earlier approaches~\citep{vae,gan} as the dominant paradigm in image generation, achieving high-quality and diverse synthesis by reversing a gradual noising process~\citep{dhariwal2021diffusionbeatgans,qi2022high,glide,imagen,ldm,dalle2,qi2024unigs}.
Among these, the Latent Diffusion Model (LDM)~\citep{ldm} \ql{wrapped with UNet structure} introduced denoising in the latent space, enabling scalable high-resolution generation~\citep{sdxl}.
More recently, transformer-based architectures~\citep{dit,transformer} have been adopted using explicit positional encoding and attention operation to further improve performance~\citep{flux,sd3,repa,sit}, and are emerging as a new paradigm for better scalability and stronger results.
\ql{We note that both UNet- and transformer-based structures benefit from the large-scale perspective datasets. Motivated by this, we leverage perspective data to compensate for the limited scale of panoramic data by inter-domain transformation and projection.}
%
% Building on this trend, we adopt the DiT architecture, which allows us to scale up generation to higher resolutions and leads to improved geometric fidelity in our results.
% \lxt{Miss cite recent SOTA DiT works, such as SORA, REPA, SiT, or related reports.}
%
% \lxt{No mention the position of our work.}

% \subsection{Panoramic image Generation}

\noindent
\textbf{Panoramic image Generation.}
Early panoramic image generation mainly relied on outpainting-based methods
~\citep{akimoto2022diverse,dastjerdi2022guided,wang2022stylelight,wang2023360,wu2023ipo,wu2023panodiffusion,lu2024autoregressive,qi2022open}, 
which reconstruct a full 360° view from partial observations, such as narrow field of view (NFoV), but often suffer from limited flexibility and content diversity. 
With advances in text-to-image generation, research has shifted towards text-to-panorama generation for more controllable results and can be broadly divided into two categories.
The first kinds of approaches~\citep{fang2023ctrl,hollein2023text2room,yu2023long,bar2023multidiffusion,lee2023syncdiffusion,li2023panogen,shi2023mvdream,mvdiffusion,park2025spherediff,yang2025omni} generate panoramic images by stitching multiple perspective views.
However, they often suffer from limited perceptual realism because of repeated objects and poor geometric fidelity, such as discontinuities.
To alleviate \ql{this problem}, some work~\citep{song2023roomdreamer,ye2024diffpano,huang2025dreamcube,kalischek2025cubediff} adopts cube mapping, which better aligns with the spherical geometry of panoramic images; 
yet discontinuities across cube faces remain unresolved, along with additional computational and temporal overhead.
Another line of work 
% \lxt{The logic here is not smooth, why call another line of the work. You can divide the method in two or three groups at the first sentence of this paragraph.}
~\citep{chen2022text2light,shum2023conditional,zhang2023diffcollage,feng2023diffusion360,ai2024dream360,wang2024customizing,yang2024dreamspace,panfusion,worldgen,smgd,hunyuanworld,ni2025makes,par,matrix3d} 
trains models directly on equirectangular images, preserving global continuity and allowing the model to learn distortion patterns. 
However, these methods struggle to maintain boundary consistency \ql{that requires seamless alignment at the 0°/360° longitude} and degrade in regions with severe polar distortion, leading to reduced geometric fidelity.
Although recent works~\citep{smgd,park2025spherediff,panfusion} attempt to alleviate these issues through alternative convolutional designs, they remain limited in practice and are less compatible with pre-trained models.
In addition, all these methods are constrained by the limited quality of panoramic datasets, often inheriting polar degradation and producing rendered-like appearances that lack perceptual realism.
In contrast, we employ a hybrid training strategy that enables the generation of high-resolution panoramic images with high perceptual realism, producing sharp and detailed content~\citep{qi2021multi} and strong geometric fidelity, ensuring correct polar distortion and seamless boundaries.
% \lxt{What are the advantages of our works.}

\section{Method}

\begin{figure}[t]
    \centering
    \includegraphics[width=\linewidth]{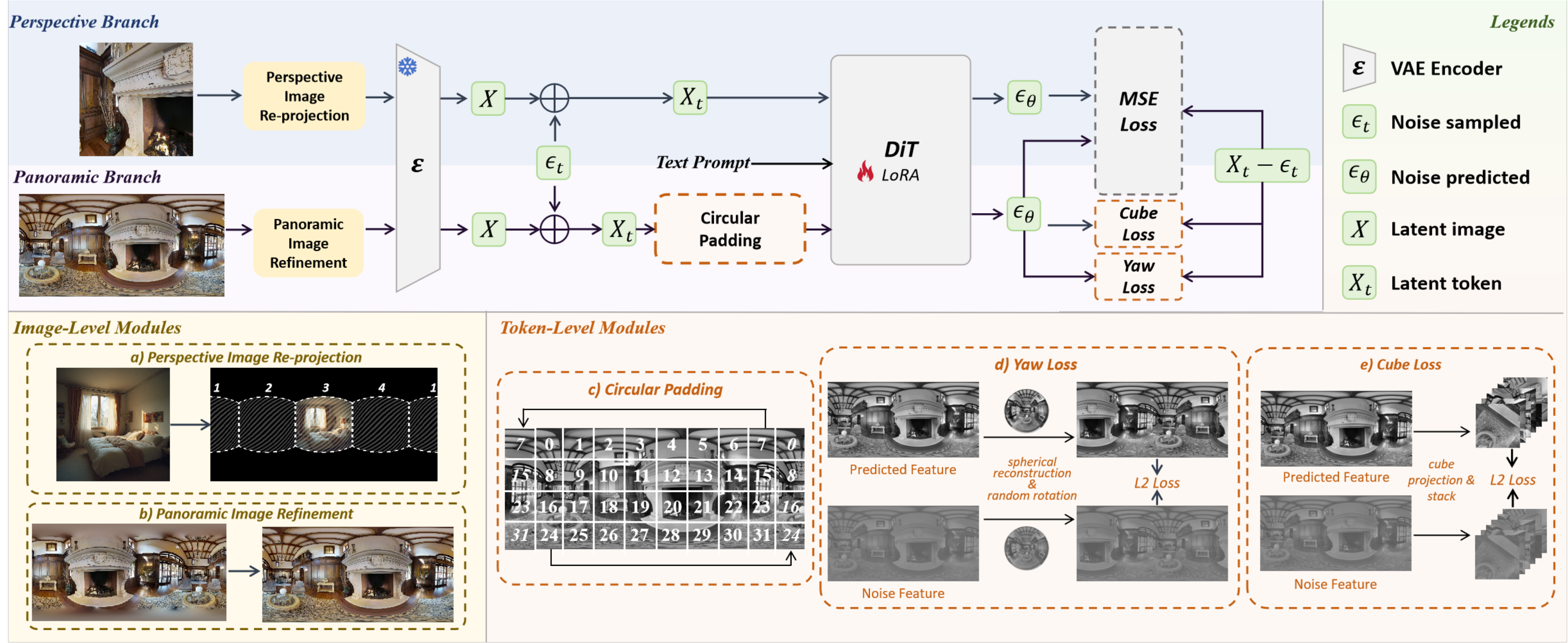}
    \caption{Overview of the \textit{DiT360} hybrid training pipeline. 
    For the perspective branch, we employ (a) perspective image re-projection to transfer perspective knowledge to panoramic domain.
    For the panoramic branch, we first apply (b) panoramic refinement to remove polar blurring and then introduce (c) position-aware circular padding, (d) rotation-consistent yaw loss and (e) distortion-aware cube loss for token-level hybrid supervision.
    }
    \label{fig:pipeline}
\end{figure}

As illustrated in~\cref{fig:pipeline}, \textit{DiT360} is a novel framework for generating panoramic images, which improves photorealism and geometric fidelity through hybrid training at both the image and token levels.
In the following sections,
We first present the preliminaries and overall design of \textit{DiT360} in \cref{sec:revisit_dit}.
We then introduce several key modules of the hybrid paradigm from two complementary perspectives: image-level regularization in \cref{sec:image_level_regularization}, and token-level supervision in \cref{sec:token_level_supervision}. Finally, we show that \textit{DiT360} natively supports extended generation tasks such as inpainting and outpainting without additional training, as detailed in \cref{sec:more_application}.

\subsection{DiT360}
\label{sec:revisit_dit}
\paragraph{Revisit Diffusion Transformer (DiT).}
Recent diffusion models increasingly adopt the DiT architecture~\citep{dit}, which uses a transformer~\citep{vit} to process latent sequences of post-VAE image tokens \(X \in \mathbb{R}^{N \times d}\), where \(N\) is the sequence length and \(d\) denotes the embedding dimension.
To capture spatial structure, DiT employs Rotary Positional Embeddings (RoPE)~\citep{su2024rope}, which inject coordinate-dependent rotations into the image tokens, thereby allowing the model to effectively encode both relative and absolute positional information.
In addition, DiT adopts a flow-based scheduler to progressively denoise the latent representation, typically conditioned on a text promp \(c\).
Its training objective is the standard denoising score-matching loss, computed as:
\begin{equation}
\mathcal{L} = \mathbb{E}_{X, c, \epsilon, t} \big[ \| \epsilon - \epsilon_\theta(X_t, c, t) \|_2^2 \big],
\end{equation}
where \(X_t\) denotes the noise latent in timestep \(t\), \(\epsilon\) is the added Gaussian noise and \(\epsilon_\theta\) represents the predicted noise of the model.

\paragraph{Overview of \textit{DiT360}.}
Building upon \dz{DiT}, we introduce \textit{DiT360} for panoramic image generation.
\dz{\cref{fig:pipeline} illustrates the proposed framework, which adopts a hybrid paradigm to jointly exploit perspective and panoramic data through two training branches. The key modules enabling hybrid training are categorized into \textbf{image-level regularization} and \textbf{token-level supervision}. At the image level, perspective image guidance and panoramic refinement introduce cross-domain knowledge to enhance perceptual quality while regularizing diversity and photorealism. At the token level, hybrid supervision across multiple objectives is conducted, which includes circular padding for boundary continuity, yaw loss for rotational robustness, and cube loss for distortion awareness. Together, this hybrid design operates across multiple representation levels to achieve perceptual photorealism and geometric fidelity.}

\subsection{\ql{Image-level Regularization}}
\label{sec:image_level_regularization}

At the image level, we adopt a hybrid regularization strategy that improves \dz{generation quality} through inter-domain transformation, combining refinement of existing panoramas with the transfer of photorealistic knowledge from perspective views.
% For image generation models, the quality of training data strongly influences the fidelity of the generated results.
% %
% However, the widely used Matterport3D dataset~\citep{Matterport3D} suffers from a severe limitation: the top and bottom regions of its equirectangular projection (ERP) images are heavily blurred, which significantly degrades the quality of the generated panoramic images. 
% %
% % Facing this problem, some prior works
% % ~\citep{hunyuanworld,layerpano3d,Structured3D} 
% % try to construct large-scale datasets using simulated environments, but the synthetic images often lack photorealism, creating a noticeable gap from real-world data.
% To tackle these challenges, we propose a hybrid dataset that combines refined panoramic images with realistic perspective images.
% %
% % Firstly, to resolve the blurring in the polar regions of ERP images, 
% % we convert them into cubemap representations, 
% % apply inpainting~\citep{flux1kontext} to the top and bottom faces with blurred areas to restore sharp details, and then convert them back into ERP format.

\begin{figure}[t]
    \centering
    \includegraphics[width=\linewidth]{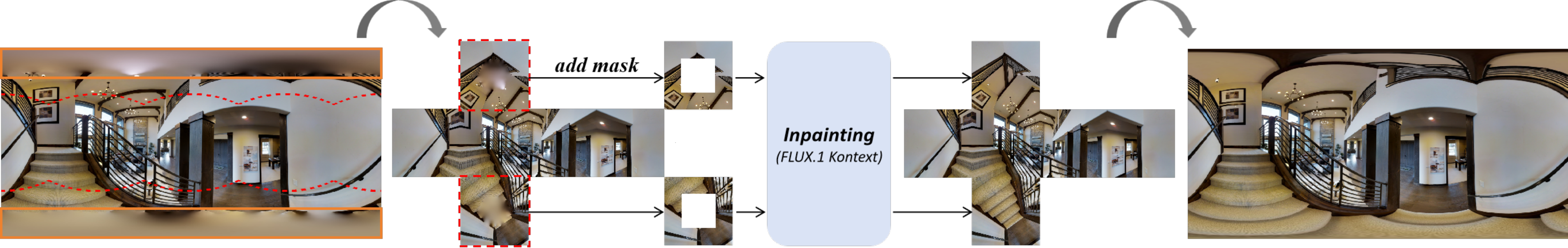}
    \caption{Panoramic image refinement pipeline. The ERP panorama is converted into a cubemap, where pre-defined masks are applied to the central regions of the top and bottom faces. These masked regions are then reconstructed with an inpainting model and reprojected to ERP. In the figure, orange boxes represent blurry areas, and red dashed boxes indicate inpainted cubes.}
    \label{fig:data_clean}
\end{figure}

\paragraph{Panoramic image refinement.} 
% To mitigate the blurring artifacts in the polar regions of ERP images, we transform the panorama into a cubemap representation. 
% %
% For each generated cubemap face $I$, we define a binary mask $M$ to localize the area for inpainting. 
% \cmt{[dz:add comment on popular panorama dataset and the highlight the artifact in fig2]} 
The availability of high-quality, real-world panoramic datasets remains severely restricted, 
with Matterport3D~\citep{Matterport3D} being one of the most widely adopted due to its large scale and high fidelity. 
Nevertheless, images in this dataset frequently exhibit blurring in the polar regions, as shown in ~\cref{fig:data_clean}, which in turn hampers the quality of downstream panoramic image generation.
\dz{To mitigate blurring artifacts in the polar regions, we transform panoramic ERPs into cubemap representations, where well-established perspective-domain inpainting can be directly applied. We first fix a binary mask $M$ for each blurred cube face $I$ to localize the inpainting area.}
Specifically, for $H=W=1024$, we mask out the central region:

\begin{equation}
M(u,v) = 
\begin{cases}
0, & \text{if } 256 \leq u,v < 768, \\
1, & \text{otherwise},
\end{cases}
\end{equation}

where $(u,v)$ denotes pixel coordinates. 
The masked image is then obtained as
\begin{equation}
I_{mask} = I \odot M + (1-M)\cdot I_{miss},
\end{equation}

where $\odot$ denotes element-wise multiplication and $I_{miss}$ is a white image of the same resolution. 
Finally, the inpainting model~\citep{flux1kontext} is then applied to $I_{mask}$ \dz{to reconstruct the missing region and produce $\hat{I}$, which is then transformed back into the ERP space to obtain blur-free, high-fidelity panoramas.}.
This process \dz{serves as an image-quality regularization step}, yielding clearer training images while \dz{constraining panoramas to retain inherent distortion characteristics}, as illustrated in \cref{fig:data_clean}.

% Secondly, to further enhance the data diversity and improve model generalization, we leverage realistic perspective images as \ql{a} supervision for panoramic generation. 
% %
% Specifically, a perspective image is randomly projected onto \ql{a} cubemap face and converted into ERP with a corresponding mask, as shown in \cref{fig:pipeline}.
%
% For both data types, captions are provided using Janus~\citep{janus}, ensuring consistent and high-quality textual supervision.
%
% Secondly, to further enhance data diversity and improve model generalization, we leverage realistic perspective images as supervision for panoramic generation. 
\paragraph{Perspective image guidance.}
In addition, we leverage high-quality realistic perspective images from the Internet
% \cmt{[dz:from which dataset?]} 
to \dz{regularize the panoramic domain by transferring photorealistic knowledge}.
Specifically, as illustrated in \cref{fig:pipeline}a, a perspective image is \dz{regarded as a cubemap lateral face and} then converted back into the ERP \dz{representation} with a corresponding mask $\mathbf{M}$. 
We \dz{restrict the re-projection to the} lateral faces, as the top and bottom faces \dz{usually correspond to} sky or ground regions, which require perspective images from \dz{uncommon viewing} angles that are \dz{rarely covered} in the dataset. 
% \cmt{[dz: need description about the training procedure using this reprojected representation]}
%
\dz{During training, we directly apply the mean squre error (MSE) loss from Flux~\citep{flux} to the re-projected ERP, restricting it to the masked regions to avoid contamination from unrelated panoramic areas, yielding:
\begin{equation}
\mathcal{L}_\text{perspective} = \mathcal{L}_\text{MSE}(\boldsymbol{\epsilon} \odot \mathbf{M}, \hat{\boldsymbol{\epsilon}_{\theta}} \odot \mathbf{M}),
\end{equation}
where $\boldsymbol{\epsilon}$ and $\hat{\boldsymbol{\epsilon}}_{\theta}$ 
denote the sampled noise and the reparameterized predicted noise, respectively.}
\dz{This strategy provides effective image-level guidance through cross-domain knowledge adaptation, exposing the model to more diverse scenes and thereby increasing the generation diversity. More importantly, the incorporated perspective knowledge regularizes the model toward photorealistic fidelity, which remains underexplored in prior works.}
% The effectiveness is validated in~\cref{fig:ablation}.

% \begin{figure}[htbp]
%     \centering
%     \includegraphics[width=\linewidth]{figure/data_clean.pdf}
%     \caption{Data preprocessing results. Left: original ERP image with distorted pole regions. Right: refined image after inpainting, where polar details are restored.}
%     \label{fig:data_clean}
% \end{figure}
% \begin{figure}[t]
%     \centering
%     \includegraphics[width=\linewidth]{figure/data_clean_pipe.pdf}
%     \caption{Panoramic image preprocessing pipeline. The ERP panorama is first converted into a cubemap representation, where binary masks are applied to the top and bottom faces to remove the central regions, later reconstructed with an inpainting model. Finally, the restored cubemap faces are projected back to ERP, yielding sharper polar textures.}
%     \label{fig:data_clean}
% \end{figure}

% \begin{figure}[t]
%     \centering
%     \includegraphics[width=\linewidth]{figure/pipeline.pdf}
%     \caption{Overview of the \textit{DiT360} training pipeline. 
%     For perspective data, the input is first mapped to a random panoramic region, with masking applied to valid areas to enable effective learning of perspective-to-panorama distortions.
%     %
%     For panoramic data, we introduce token-level circular padding with shared position embedding to ensure boundary coherence and incorporate cube and yaw loss to improve detail preservation and rotational consistency. 
%     }
%     \label{fig:pipeline}
% \end{figure}

% \subsection{DiT360}
\subsection{\ql{Token-level Supervision}}
\label{sec:token_level_supervision}
% Our framework preserves the original architecture with minimal modifications, ensuring full compatibility with pretrained diffusion transformers.
% %
% The training pipeline is based on the hybrid dataset~\cref{sec:hybrid_training_dataset}.  
% For panoramic data, we incorporate circular padding in~\cref{sec:circular_padding} to enhance rotational consistency, and introduce yaw and cube losses in~\cref{sec:training_loss} to improve rotation alignment and capture distortions patterns, respectively.
% %
% For perspective data, we apply a masking strategy and supervise only within the masked regions, enabling effective learning of distortions from perspective to panorama.  
% %
% An overview of the training pipeline is illustrated in \cref{fig:pipeline}.  
At the token level, \textit{DiT360} \dz{adopts a hybrid training strategy that balances complementary supervision at the post-VAE token level, simultaneously enhancing boundary continuity, rotational robustness, distortion awareness and perceptual quality.}
Specifically, we introduce three mechanisms applied to noisy tokens of panoramic inputs: 
position-aware circular padding for seamless boundary coherence, 
yaw loss for \dz{global} rotation consistency, 
and cube loss for \dz{precise supervision of ERP} distortion patterns. 
\dz{Together, we propose a hybrid loss design to ensure fine-grained token-level supervision while maintaining balanced generation quality across multiple dimensions.}

\paragraph{\ql{Position-aware Circular Padding.}}
\label{sec:circular_padding}

% \lxt{I am confused on two things: 1, Why Position-aware Circular Padding is put in the hybrid training.}
% Panoramic images span a full 360° horizontal view, which poses a unique challenge of maintaining consistency at the image boundaries. 
%
% Prior studies~\citep{shum2023conditional,wang2023360,panfusion} have explored the use of circular padding to address this issue.
%
% However, since the UNet relies on convolutions to capture spatial information, circular padding alone often fails to ensure seamless boundaries.
%
% In addition, some methods~\citep{layerpano3d,hunyuanworld} adopt a cyclic denoising strategy.
% While effective in some cases, this pixel-wise blending approach often introduces a distinct seam in the image.
%
Panoramic images cover the full 360° horizontal field, making it critical to maintain continuity across image boundaries.
%
% To address this challenge, we propose a token-level circular padding mechanism tailored for diffusion transformers (DiT)~\citep{dit}.
% %
% Since positional encodings in DiTs are inherently aligned with image content~\citep{pa},
% we extend this property to latent tokens, enforcing spatial consistency and enhancing boundary coherence without architectural changes.
%
To address this challenge,
we propose a token\dz{-based} circular padding mechanism specific to \dz{DiT-based frameworks} that takes advantage of the inherent correspondence between explicit positional encoding and image content.
This property ensures that latent tokens at the same spatial position generate consistent visual features, which we exploit to enhance boundary coherence without introducing additional architectural complexity.
%
% Specifically, as illustrated in \cref{fig:pipeline}b, after the VAE compression and the subsequent noise injection, we reshape the latent tokens \(X_t \in \mathbb{R}^{N \times d}\) into \(\mathbb{R}^{H \times W \times d}\), and extract the first and last column features \(X_{0}\) and \(X_{-1}\). 
Specifically, as illustrated in \cref{fig:pipeline}c, 
after the VAE compression and the subsequent noise injection, 
we reshape the latent tokens \(X_t \in \mathbb{R}^{N \times d}\) 
into \(X_t \in \mathbb{R}^{H \times W \times d}\). 
We then apply a circular padding along the width dimension.
Formally, let \(X_{0}\) and \(X_{-1}\) denote the first and last column features, respectively. 
We then concatenate them to obtain the padded tensor
\begin{equation}
\tilde{X}_t 
= \big[\, X_{-1},\; X_t,\; X_{0} \,\big]
\;\in\; \mathbb{R}^{H \times (W+2) \times d}.
\end{equation}
The same operation is applied to the positional encoding,
which encourages the model to learn continuity specifically 
between adjacent columns across the boundary.
\paragraph{Rotation-consistent Yaw Loss.}
To enforce global rotational robustness, we introduce yaw loss that \dz{offers token-level supervision on} the model’s behavior under spherical rotations along the yaw axis, as illustrated in \cref{fig:pipeline}d. 
Unlike the standard diffusion loss, \dz{the corresponding} yaw loss captures the non-linear effects of yaw rotations and constrains the model to produce consistent predictions across different \dz{viewing angles}.
Formally, with the reparameterized noise $\boldsymbol{\epsilon}$ and predicted noise $\boldsymbol{\epsilon}_\theta$, we randomly select a yaw rotation angle $a$ and define the rotated features as:
\begin{equation}
\boldsymbol{\epsilon}_\text{yaw} = \text{Rotate}(X_{t}-\boldsymbol{\epsilon}, a),
\quad
\boldsymbol{\epsilon}_{\theta,\text{yaw}} = \text{Rotate}(\boldsymbol{\epsilon}_\theta, a),
\end{equation}
where $\text{Rotate}(\cdot, a)$ denotes the equirectangular panorama rotated by angle $a$ along the yaw axis.

The yaw loss is then computed as the mean squared error (MSE) between the predicted and target rotated noise features:
\begin{equation}
\mathcal{L}_\text{yaw} =
\mathbb{E} \left[
\left| \boldsymbol{\epsilon}_{\theta,\text{yaw}} - \boldsymbol{\epsilon}_\text{yaw} \right|_2^2
\right].
\end{equation}
% Further evidence of its effectiveness is provided in~\cref{fig:ablation}.

% \paragraph{Cube Loss.} 
\paragraph{Distortion-aware Cube Loss.}
% To better capture distortion patterns and enhance fine-grained details, we introduce a cube loss based on the cubemap representation of equirectangular panoramas. 
% %
% The core idea is to project both the sampled noise and the model-predicted noise onto cube faces and supervise them in a face-wise manner, thereby leveraging the model’s strengths in perspective image generation.
To effectively model distortion patterns and preserve fine details, we introduce a cube loss based on the cubemap representation of panoramasas, as shown in \cref{fig:pipeline}e. 
Direct supervision on equirectangular projections often causes the model to reproduce similar distorted appearances rather than learn the precise structural patterns, which leads to incorrect generation details with dealing with polar-region distortions.
\dz{To address this challenge, we project both sampled and predicted noise onto cube faces and apply face-wise supervision, thereby transferring model's strength in perspective priors to the panoramic domain to preserve structural distortion patterns.}
Further analysis are provided in~\cref{appendix:polar_distortions}.
Formally, let $\mathbf{X}_{t}$ denote the noisy latent at time step $t$ in the forward diffusion process, $\boldsymbol{\epsilon}$ denote the reparameterized Gaussian noise, and $\boldsymbol{\epsilon}_\theta$ denote the noise predicted by the model. 
We define the cube-space features by applying a cube-mapping operation:
\begin{equation}
\boldsymbol{\epsilon}_\text{cube} = \text{CubeMap}(X_{t}-\boldsymbol{\epsilon}),
\quad
\boldsymbol{\epsilon}_{\theta,\text{cube}} = \text{CubeMap}(\boldsymbol{\epsilon}_\theta),
\end{equation}
where $\text{CubeMap}(\cdot)$ transforms an equirectangular panorama into six cube faces. 
Then, the cube loss is computed as the MSE between the predicted and target cube-space noise features: 
\begin{equation}
\mathcal{L}_\text{cube} =
\mathbb{E} \left[
\left| \boldsymbol{\epsilon}_{\theta,\text{cube}} - \boldsymbol{\epsilon}_\text{cube} \right|_2^2
\right].
\end{equation}
% Additional validation can be found in~\cref{fig:ablation} and ~\cref{appendix:polar_distortions}.
It is worth noting that we apply both cube and yaw losses directly in the latent noise space. 
While a natural alternative is to compute them in the latent token space—by predicting latents from noise and comparing with ground-truth latents—our experiments show that noise predictions already encode rich spatial and structural information due to the coupling of noise and semantics in the diffusion objective, making them suitable for spatial supervision. 
In addition, operating in the noise space aligns the auxiliary losses with the flow-based scheduler, thereby improving training stability.

\paragraph{Hybrid Loss Design.}
\dz{To better balance geometric fidelity and perceptual quality, we adopt a hybrid loss design that retains the MSE loss from Flux~\citep{flux} as the principal objective and augment it with yaw loss and cube loss described above. The overall training loss $\mathcal{L}_\text{pano}$ for the panoramic branch is then calculated as: \begin{equation}
\mathcal{L}_\text{pano} = \mathcal{L}_\text{MSE} + \lambda_{1}\mathcal{L}_\text{cube} + \lambda_{2}\mathcal{L}_\text{yaw},
\end{equation}
where $\lambda{1}$ and $\lambda_{2}$ represent the balancing coefficients. This token-level hybrid supervision ensures that general perceptual quality, rotational robustness and distortion fidelity are jointly preserved within a unified training framework.}
%To enhance both the diversity of training data and the fidelity of generated textures, we adopt a hybrid training strategy based on the dataset described in~\cref{sec:image_level_regularization}.  
%
%We take the mean squared error (MSE) loss from the original Flux~\citep{flux} training as our baseline objective, and extend it with additional constraints tailored to each data modality.  

%For panoramic images, two complementary objectives are introduced: the cube loss and the yaw loss.
%
%The cube loss supervises fine-grained detail reconstruction by projecting latents into cube maps, while the yaw loss enforces rotational consistency under spherical transformations.
%

\subsection{More Applications}
\label{sec:more_application}

Benefiting from the strong robustness of our method, we perform feature replacement via inversion ~\citep{pa} to enable image inpainting and outpainting without additional training.  
In addition, our model natively supports high-resolution generation, with all training conducted at 1024$\times$2048 resolution.
The results in \cref{fig:teaser} \dz{demonstrate} the generalization capability of our approach beyond the primary generation task, with
more analysis and results provided in ~\cref{appendix:inpainting_and_outpainting}.
\section{Experiments}

\begin{figure}[t]
    \centering
    \includegraphics[width=\linewidth]{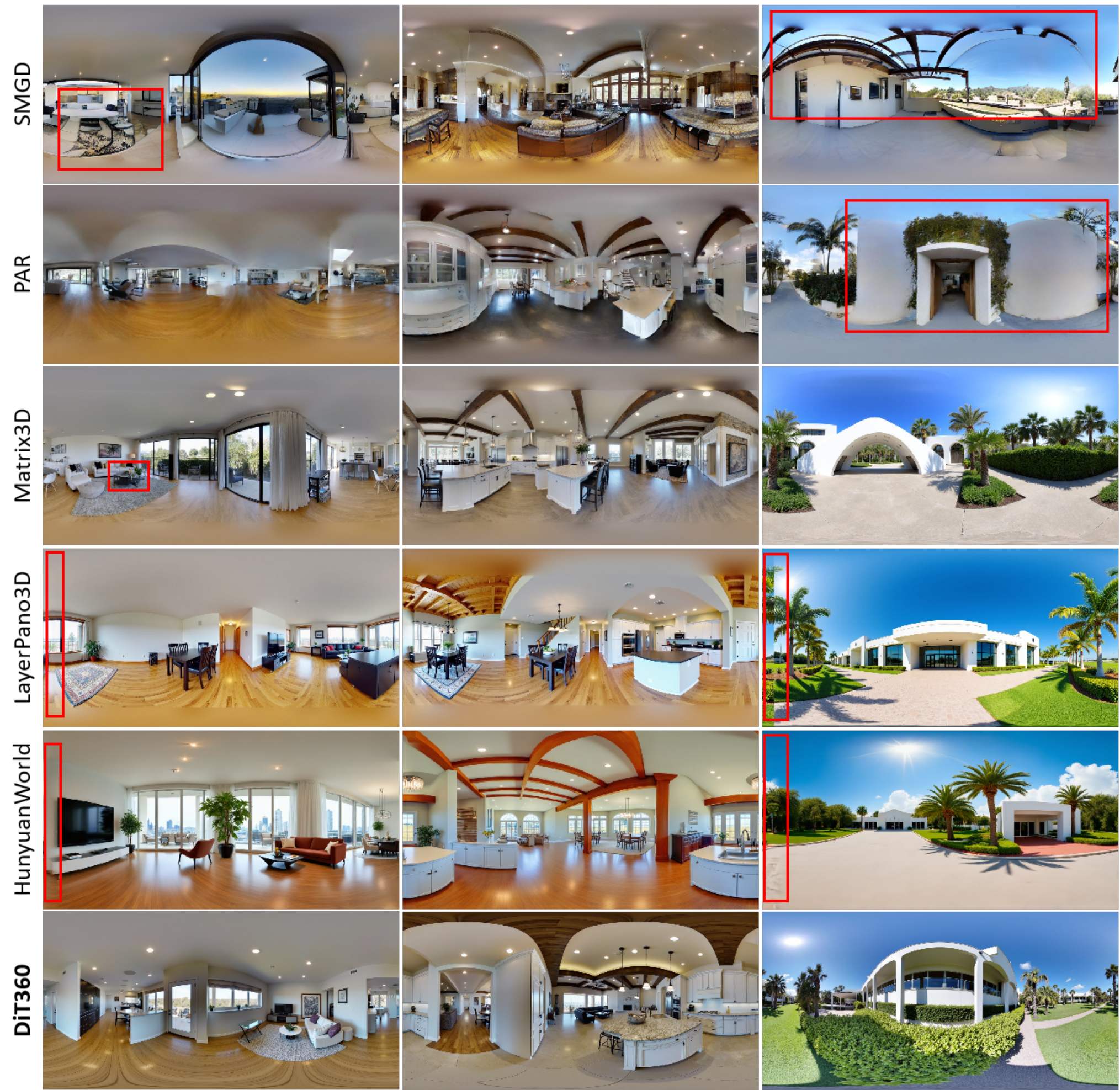}
    \caption{
    Qualitative comparisons on panorama generation. The representative artifacts are highlighted with red boxes. More complete results are provided in \cref{appendix:full_comparision}.
    }
    \label{fig:comparison}
\end{figure}

\subsection{Setup}
\label{sec:setup}

\textit{DiT360} is developed on top of Flux~\citep{flux} with LoRA~\citep{lora} incorporated into the attention layers.
For improved perceptual realism and geometric fidelity, we design a hybrid training strategy that combines high-quality Internet landscape images with large-scale panoramas from Matterport3D~\citep{Matterport3D}.
To assess the effectiveness of our approach, we adopt a diverse set of complementary metrics covering realism, diversity, text–image alignment, and perceptual quality, ensuring a comprehensive assessment of model performance. 
More detailed descriptions of the implementation, dataset preprocessing, and metric definitions are in~\cref{appendix:experiment_settings}.

\begin{figure}[t]
    \centering
    \includegraphics[width=\linewidth]{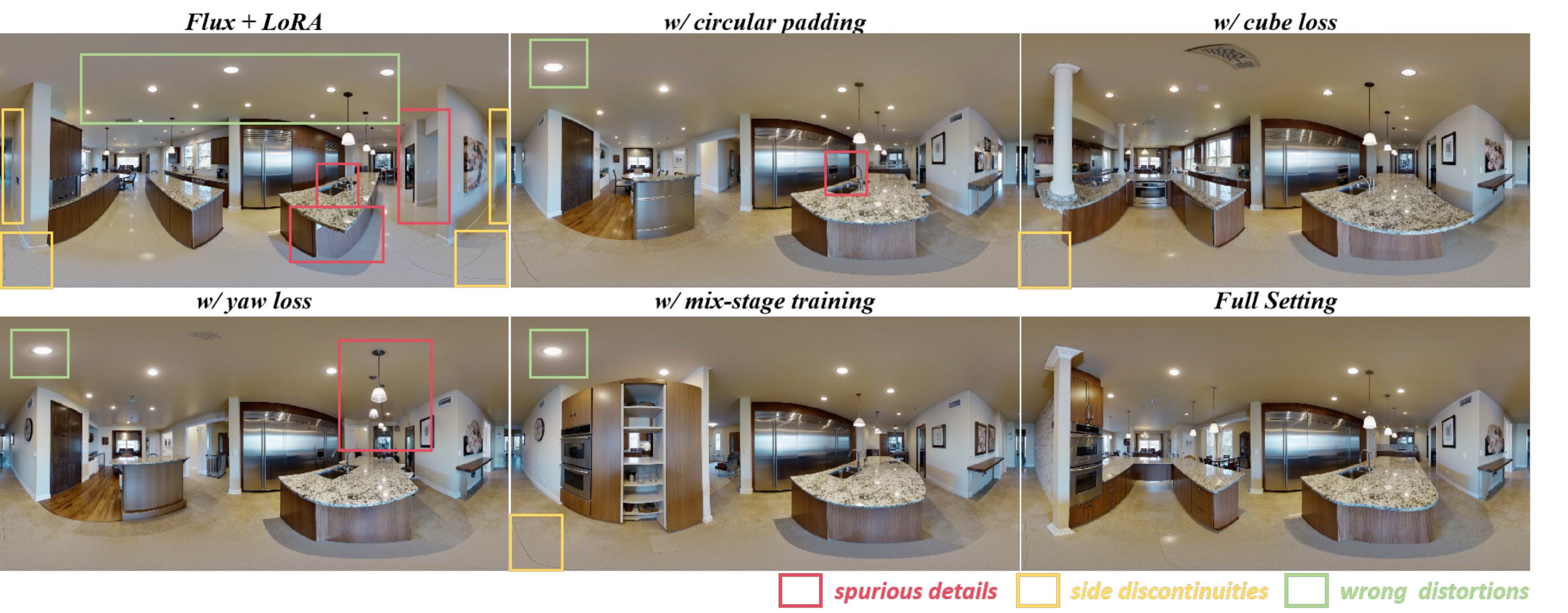}
    \caption{Ablation results of different settings. Artifacts are marked by color-coded bounding boxes: red for spurious details, yellow for boundary discontinuities, and green for incorrect distortions.}
    \label{fig:ablation}
    \vspace{-0.2in}
\end{figure}

\begin{table}[t]
\centering
\caption{Quantitative comparison results on text-to-panorama generation. We highlight the best result in \textbf{bold} and the second best with \underline{underline}.}
\label{tab:metrics}
\setlength{\tabcolsep}{3pt}
\resizebox{\textwidth}{!}{
\begin{tabular}{lccccccccccc} 
\toprule
Methods & 
FID$\downarrow$ & 
FID\textsubscript{clip}$\downarrow$ & 
FID\textsubscript{pole}$\downarrow$ & 
FID\textsubscript{equ}$\downarrow$ & 
FAED$\downarrow$ & 
IS$\uparrow$ & 
CS$\uparrow$ & 
QA\textsubscript{quality}$\uparrow$ &
QA\textsubscript{aesthetic}$\uparrow$ &
BRISQUE$\downarrow$ &
NIQE$\downarrow$ \\
\midrule
PanFusion       & 124.87 & 120.75 & 182.09 & 108.12 & 11.06 & 1.30 & 28.35 & 3.83 & 3.56 & 27.38 & 4.31 \\
MVDiffusion     & 108.19 & 117.26 & - & - & 4.39 & \underline{1.58} & 34.65 & 3.97 & 3.25 & 44.79 & 4.91 \\
SMGD            & \underline{46.72} & \underline{45.04} & \underline{65.69} & 34.84 & 3.29 & 1.40 & 31.14 & 4.05 & 3.77 & 30.35 & 4.75 \\
PAR             & 47.72 & 47.26 & 76.93 & 27.39 & \underline{2.97} & 1.34 & 33.85 & 3.91 & 3.54 & 32.26 & 4.38 \\
WorldGen        & 67.11 & 62.97 & 79.32 & 33.45 & 3.29 & 1.40 & 34.61 & 4.30 & 3.59 & 32.31 & 4.82 \\
Matrix-3D       & 60.91 & 56.70 & 77.21 & \underline{26.73} & 3.08 & 1.56 & 34.59 & 4.48 & 3.78 & \underline{16.37} & 3.95 \\
LayerPano3D     & 62.82 & 60.34 & 80.37 & 38.67 & 2.98 & 1.50 & 34.40 & \textbf{4.73} & \underline{3.93} & 33.91 & \underline{3.79} \\
HunyuanWorld    & 76.75 & 75.65 & 106.58 & 41.75 & \textbf{2.91} & 1.53 & \textbf{34.73} & 4.67 & 3.85 & 39.12 & 5.18 \\
\midrule
\textbf{Ours}   & \textbf{42.88} & \textbf{41.60} & \textbf{50.88} & \textbf{24.77} & \textbf{2.91} & \textbf{1.60} & \underline{34.68} & \underline{4.69} & \textbf{4.19} & \textbf{10.25} & \textbf{3.72} \\
% \midrule
% Flux + LoRA            & 46.69 & 45.90 & 66.03 & 28.91 & 3.23 & 1.51 & 34.39 & 4.40 & 3.97 & 17.02 & 3.97 \\
% w/ circular padding    & 43.71 & 42.36 & 61.32 & 27.51 & 3.04 & 1.54 & 34.44 & 4.51 & 3.98 & 13.61 & 3.82 \\
% w/ cube loss           & 44.40 & 43.75 & 60.16 & 26.30 & 3.01 & 1.57 & 34.62 & 4.41 & 3.92 & 15.68 & 3.89 \\
% w/ yaw loss            & 44.63 & 43.90 & 64.19 & 26.99 & 2.98 & 1.56 & 34.53 & 4.37 & 3.94 & 15.96 & 3.92 \\
% w/ mix-stage train.    & 46.03 & 44.92 & 63.72 & 27.81 & 2.95 & 1.48 & 34.42 & 4.54 & 4.02 & 16.94 & 3.83 \\
\bottomrule
\end{tabular}
}
\vspace{-0.2in}
\end{table}

\subsection{Main Results and Comparisons}

In this section, we present our main experimental results and conduct a comprehensive comparison with representative baselines, including PanFusion~\citep{panfusion}, MVDiffusion~\citep{mvdiffusion}, SMGD~\citep{smgd}, PAR~\citep{par}, WorldGen~\citep{worldgen}, Matrix-3D~\citep{matrix3d}, LayerPano3D~\citep{layerpano3d}, and HunyuanWorld~\citep{hunyuanworld}. 
These methods span diverse architectures, enabling a comprehensive evaluation of our approach. 
In addition to quantitative and qualitative comparisons, we also present a user study in~\cref{appendix:user_study} and further results in~\cref{appendix:more_results}.

\paragraph{Qualitative Comparisons.}

We provide qualitative comparisons with baseline methods in \cref{fig:comparison} and highlights artifacts with red boxes. 
%
% MVDiffusion~\citep{mvdiffusion} synthesizes pseudo-panoramas by stitching multi-view inputs but introduces redundant elements. 
%
% PanFusion~\citep{panfusion}, 
SMGD~\citep{smgd} and PAR~\citep{par} propose alternative paradigms—structural modifications or autoregression—but struggle with detail fidelity, often producing cluttered or less precise results.
Moreover, insufficient data quality leads to pronounced distortions near the polar regions, resulting in poor perceptual realism.
Recent advances in Diffusion Transformers (DiT)~\citep{dit} have led to their adoption as backbones in several panorama generation methods.
%
% WorldGen~\citep{worldgen} achieves relatively high quality but fails to ensure lateral continuity, 
%
Matrix-3D improves boundary alignment yet struggles with fine-grained details,
suffering from limited geometric fidelity.
LayerPano3D~\citep{layerpano3d} and HunyuanWorld~\citep{hunyuanworld} leverage large amounts of synthetic data, which improves geometric fidelity to some extent, but results in render-like appearances that compromise perceptual realism; additionally, iterative denoising introduces further artifacts.
In contrast, our method generates panoramas with high perceptual realism and geometric fidelity, producing sharp, detail-preserving images with strong lateral consistency and effectively mitigated distortions.

\paragraph{Quantitative Comparisons.}

We further conduct quantitative evaluations to validate the effectiveness of our approach, with results reported in \cref{tab:metrics}. 
Our method ranks first on nearly all benchmarks and shows consistently strong performance across most metrics.
Although our approach slightly underperforms the top methods on CLIP Score and the quality branch of Q-Align, the gaps are marginal and largely attributable to the fact that both metrics are designed for perspective images, which may not fully reflect the quality and fidelity of panoramas.
%
% Interestingly, some DiT-based methods~\citep{worldgen,hunyuanworld,layerpano3d,matrix3d} produce visually superior panoramas, yet their FID scores are lower than those of approaches with less effective visual results~\citep{par,smgd}. 
%
% This discrepancy is primarily due to their reliance on synthetic training data, which introduces a mismatch when computing FID against real images.
%
Collectively, the results support our qualitative observations and demonstrate the effectiveness and robustness of our approach in generating high-quality panoramas.

% \begin{figure}[t]
%     \centering
%     \includegraphics[width=\linewidth]{figure/ablation.pdf}
%     \caption{Ablation results of different settings. Artifacts are marked by color-coded bounding boxes: red for spurious details, yellow for side discontinuities, and green for wrong distortions.}
%     \label{fig:ablation}
% \end{figure}

\begin{table}[t]
\centering
\caption{Ablation study of different model components on text-to-panorama generation.
Best results are in \textbf{bold}, second best are \underline{underlined}, and “Pi guidance” denotes perspective image guidance.
}
\label{tab:ablation}
\setlength{\tabcolsep}{3pt}
\resizebox{\textwidth}{!}{
\begin{tabular}{lccccccccccc} 
\toprule
Methods & 
FID$\downarrow$ & 
FID\textsubscript{clip}$\downarrow$ & 
FID\textsubscript{pole}$\downarrow$ & 
FID\textsubscript{equ}$\downarrow$ & 
FAED$\downarrow$ & 
IS$\uparrow$ & 
CS$\uparrow$ & 
QA\textsubscript{quality}$\uparrow$ &
QA\textsubscript{aesthetic}$\uparrow$ &
BRISQUE$\downarrow$ &
NIQE$\downarrow$ \\
\midrule
Flux + LoRA            & 46.69 & 45.90 & 66.03 & 28.91 & 3.23 & 1.51 & 34.39 & 4.40 & 3.97 & 17.02 & 3.97 \\
w/ circular padding    & \textbf{43.71} & \textbf{42.36} & \underline{61.32} & 27.51 & 3.04 & 1.54 & 34.44 & \underline{4.51} & \underline{3.98} & \textbf{13.61} & \textbf{3.82} \\
w/ cube loss           & \underline{44.40} & \underline{43.75} & \textbf{60.16} & \textbf{26.30} & 3.01 & \textbf{1.57} & \textbf{34.62} & 4.41 & 3.92 & \underline{15.68} & 3.89 \\
w/ yaw loss            & 44.63 & 43.90 & 64.19 & \underline{26.99} & \textbf{2.98} & \underline{1.56} & \underline{34.53} & 4.37 & 3.94 & 15.96 & 3.92 \\
w/ pi guidance    & 46.03 & 44.92 & 63.72 & 27.81 & \underline{2.95} & 1.48 & 34.42 & \textbf{4.54} & \textbf{4.02} & 16.94 & \underline{3.83} \\
\midrule
\textbf{Ours (w/ all)}   & 42.88 & 41.60 & 50.88 & 24.77 & 2.91 & 1.60 & 34.68 & 4.69 & 4.19 & 10.25 & 3.72 \\
\bottomrule
\end{tabular}
}
\vspace{-0.2in}
\end{table}

\subsection{Ablation Study}

To assess the contribution of each component, we conduct ablation studies using a combination of Flux~\citep{flux} and LoRA~\citep{lora} as the baseline.
We ablate four key modules: position-sensitive circular padding, distortion-sensitive cube loss, rotation-consistent yaw loss, and perspective
image guidance and evaluate their impact in~\cref{tab:ablation} and \cref{fig:ablation}.

\textbf{Circular padding} significantly enhances consistency across image boundaries and also improves overall image quality, reflected in reductions of FID~\citep{fid} and BRISQUE~\citep{brisque}, because the identical positional encoding on the left and right edges allows the model to learn correct boundary correspondences.

\textbf{Cube loss} mainly refines fine-grained details and reduces artifacts by applying additional supervision on the cubemap representation, enabling the model to learn accurate panoramic distortions.
This results in substantially fewer artifacts in the polar regions and thus largely improved IS and CS that are more related to the visual semantics.

\textbf{Yaw loss} improves global rotation consistency and structural coherence, explaining its superior performance on FAED~\citep{faed} where the autoencoders used are pre-trained by panoramic images. This is because that we supervise the model on rotated tokens to explicitly enforce full-image rotation consistency.

\textbf{Perspective image guidance} further enhances local details, enriches visual diversity and effectively mitigates detail-related artifacts, as evidenced in the QA\textsubscript{quality} and
QA\textsubscript{aesthetic} metrics that are more sensitive to the visual style.

Overall, the components contribute to the perceptual realism and geometric fidelity, and their combination delivers the strongest performance, validating the effectiveness of our framework.

\vspace{-0.1in}
\section{conclusion}
\label{sec:conclusion}

In this paper, we proposed \textit{DiT360}, a framework for geometry-aware and photorealistic panoramic image generation, built upon a hybrid training strategy that combines limited high-quality panoramic data with large-scale perspective images to enhance both realism and generalization. 
To fully leverage this hybrid paradigm, we introduce multiple modules across different representation levels, where image-level regularization refines existing panoramas and leverages perspective data to enhance diversity and photorealism, while token-level supervision in the latent space enforces geometric consistency through rotation- and distortion-aware constraints.
Extensive experiments on text-to-panorama generation, inpainting, and outpainting demonstrate superior image fidelity, boundary consistency, and visual quality. 
By bridging perspective and panoramic domains across multiple representation levels, \textit{DiT360} establishes a strong baseline for future research in 3D scene generation and large-scale open-world environments.

\bibliography{iclr2026_conference}
\bibliographystyle{iclr2026_conference}

\appendix
\section*{Appendix}
\label{sec:appendix}

\section{Effect of Supervision on Polar Distortions}
\label{appendix:polar_distortions}

\begin{figure}[htbp]
    \centering
    \includegraphics[width=\linewidth]{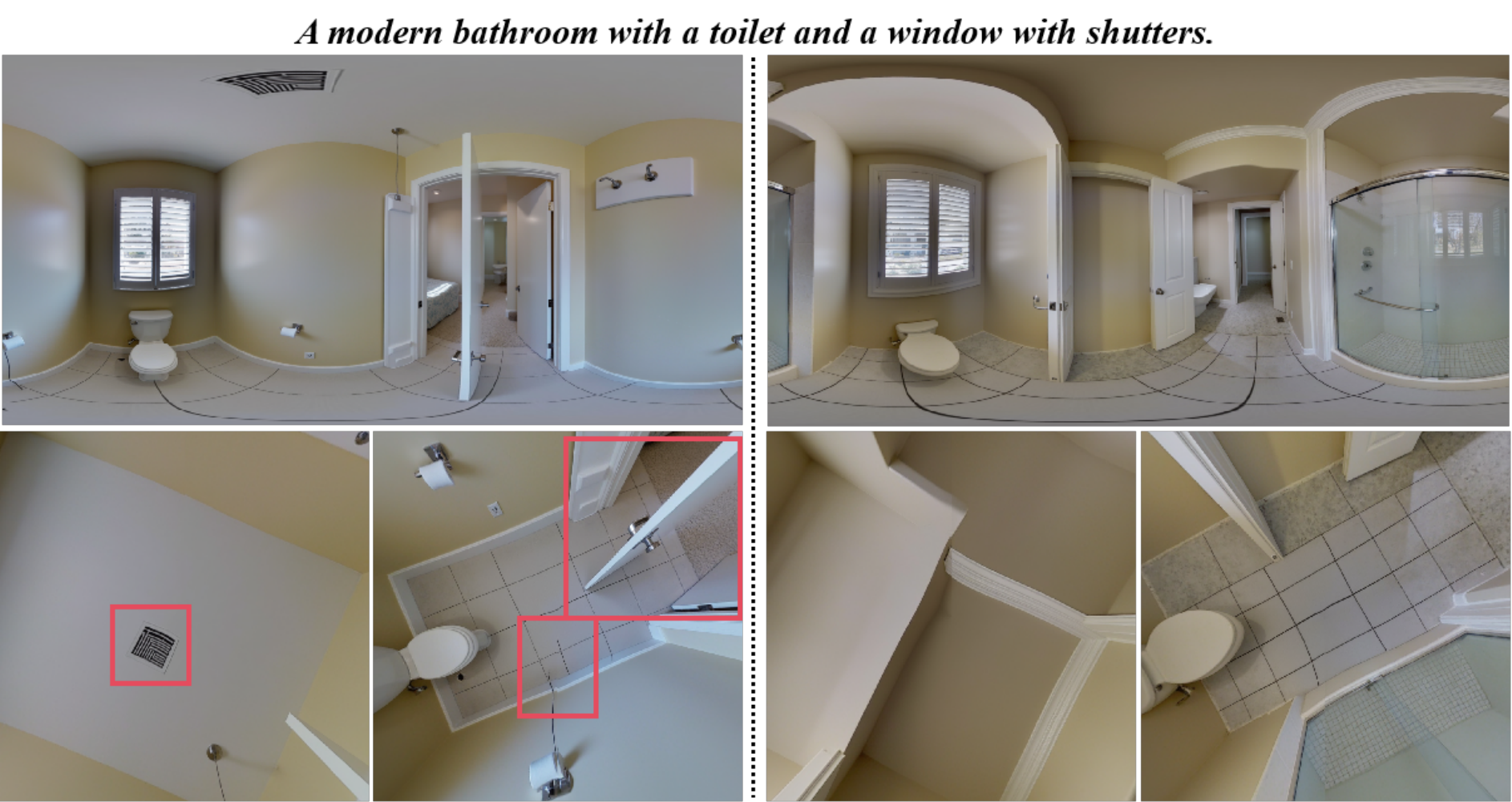}
    \caption{Qualitative comparison of generated panoramas and their top/bottom cube faces without (left) and with (right) cube loss. Red boxes mark regions where polar artifacts are significantly reduced when supervision is applied.} 
    \label{fig:comparison_cube_loss}
\end{figure}

In this section, we further illustrate the effect of cube loss in addressing severe distortions around the polar regions. 
Figure~\ref{fig:comparison_cube_loss} compares results generated from the same prompt without and with this supervision, showing that incorporating cube loss leads to clearer structures and fewer artifacts in the polar regions.

\section{Inpainting and Outpainting}
\label{appendix:inpainting_and_outpainting}

\begin{figure}[t]
    \centering
    \includegraphics[width=\linewidth]{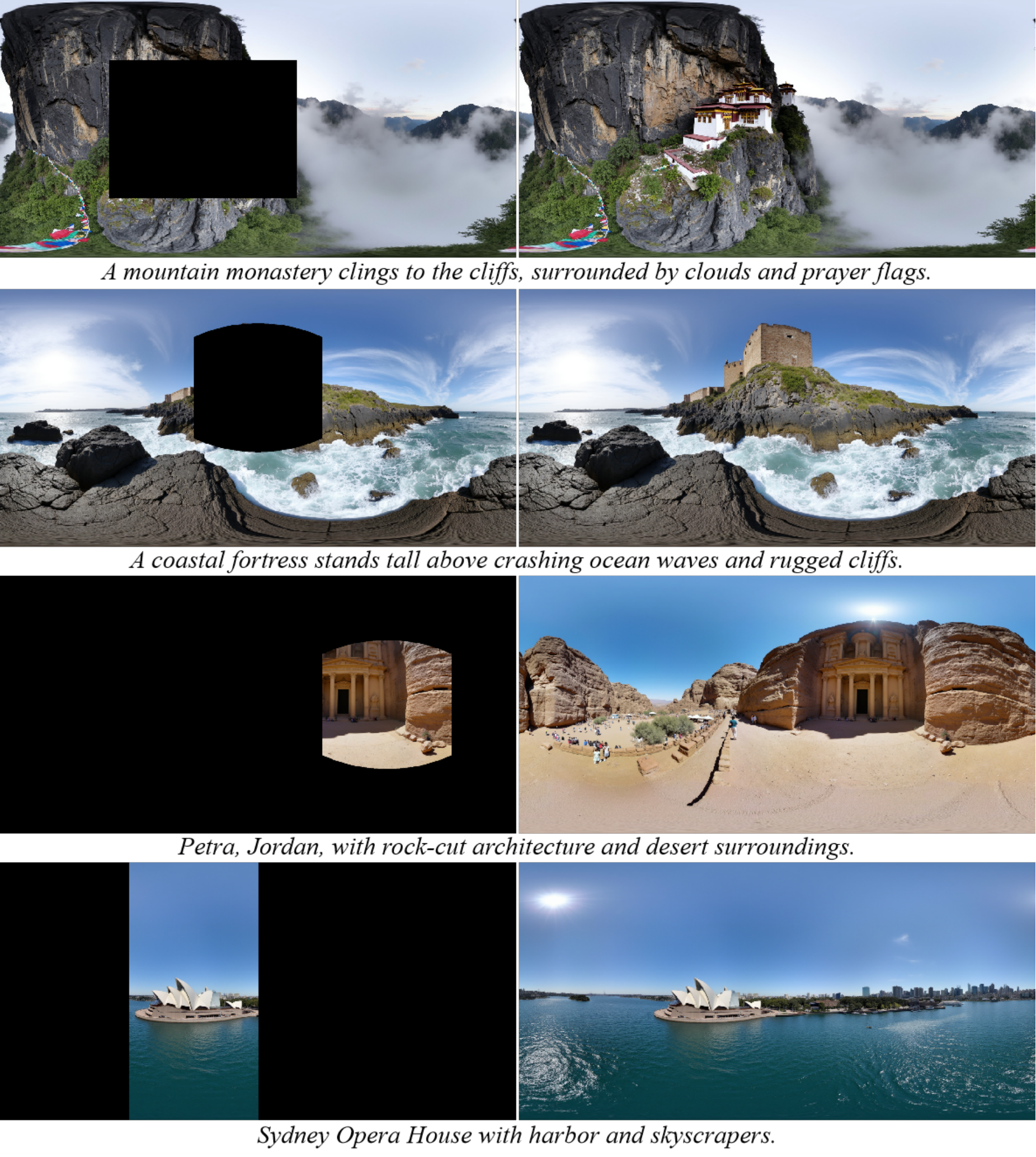}
    \caption{More results on inpainting and outpainting.} 
    \label{fig:inpainting_outpainting_results}
\end{figure}

\textit{DiT360} demonstrates native inpainting and outpainting capabilities without requiring additional training, thereby establishing a unified framework for panoramic image generation.
Specifically, inspired by ~\citep{pa}, we first perform inversion on the input image to obtain its initial noise representation. 
At the same time, we extract reference image tokens without positional encodings, along with the associated subject mask.
During the early denoising steps, we employ a token replacement strategy. 
The tokens within the masked or extended regions are substituted with those from the reference image, while preserving the original positional encodings. 
This time-step-adaptive replacement mechanism ensures faithful reproduction of subject details and spatial consistency. 
It anchors subject identity in the early phase of generation and naturally guides the model toward coherent content completion.
As a result, \textit{DiT360} produces consistent and semantically rich results in both inpainting and outpainting tasks. 
More results are provided in \cref{fig:inpainting_outpainting_results}.

\section{Experiment Settings}
\label{appendix:experiment_settings}
\paragraph{Implementation Details.}
We developed \textit{DiT360} on top of Flux~\citep{flux}, integrating LoRA~\citep{lora} into the attention layers. 
The model was fine-tuned on 5 H20 GPUs using AdamW~\citep{adamw} with a learning rate of $2\times10^{-5}$ for 20 epochs, a batch size per GPU of 1, and a gradient accumulation of 3. 
Our experiments revealed that the guidance scale plays a crucial role in convergence, with 1.0 yielding the most stable training. 
For inference, we set the guidance scale to 3.0 and employed 28 sampling steps.
 
\paragraph{Dataset.}
We adopt a hybrid training strategy that combines perspective and panoramic data. 
For the perspective branch, we curate 40k high-quality landscape images from the Internet, center-crop them to a 1:1 ratio, and project them onto random panoramic regions. 
For the panoramic branch, we follow PanFusion~\citep{panfusion} and utilize Matterport3D~\citep{Matterport3D}, a large-scale RGB-D dataset comprising 10,800 panoramas across 90 building-scale scenes. 
To mitigate distortion, we refine the blurred polar regions and use 10k panoramas for training while reserving the remainder for validation, consistent with prior work.

\paragraph{Evaluation Metrics.}
Following prior work, we evaluate our method with a diverse set of complementary metrics.
For realism, we adopt Fréchet Inception Distance (FID)~\citep{fid} and its variants, including FID\textsubscript{clip} for fair comparison by excluding blurred polar regions, and FID\textsubscript{pole} and FID\textsubscript{equ} following SMGD~\citep{smgd} to assess polar distortion and perspective projection quality. 
Since FID relies on an Inception network trained on perspective images and may not fully capture panoramic characteristics, we further employ Fréchet Auto-Encoder Distance (FAED)~\citep{faed}, a variant tailored for panoramas. 
For diversity, we report Inception Score (IS)~\citep{is}, replacing the standard Inception-v3~\citep{inceptionv3} with a ResNet pretrained on Places365~\citep{resnet,places365} to better reflect the scene-centric nature of our data.
For text–image alignment, we compute CLIP Score (CS)~\citep{clip}, and for perceptual quality, we report Q-Align (QA)~\citep{q_align}, BRISQUE~\citep{brisque}, and NIQE~\citep{niqe}, following HunyuanWorld~\citep{hunyuanworld}.

\section{Full Comparision}
\label{appendix:full_comparision}

\begin{figure}[t]
    \centering
    \includegraphics[width=\linewidth]{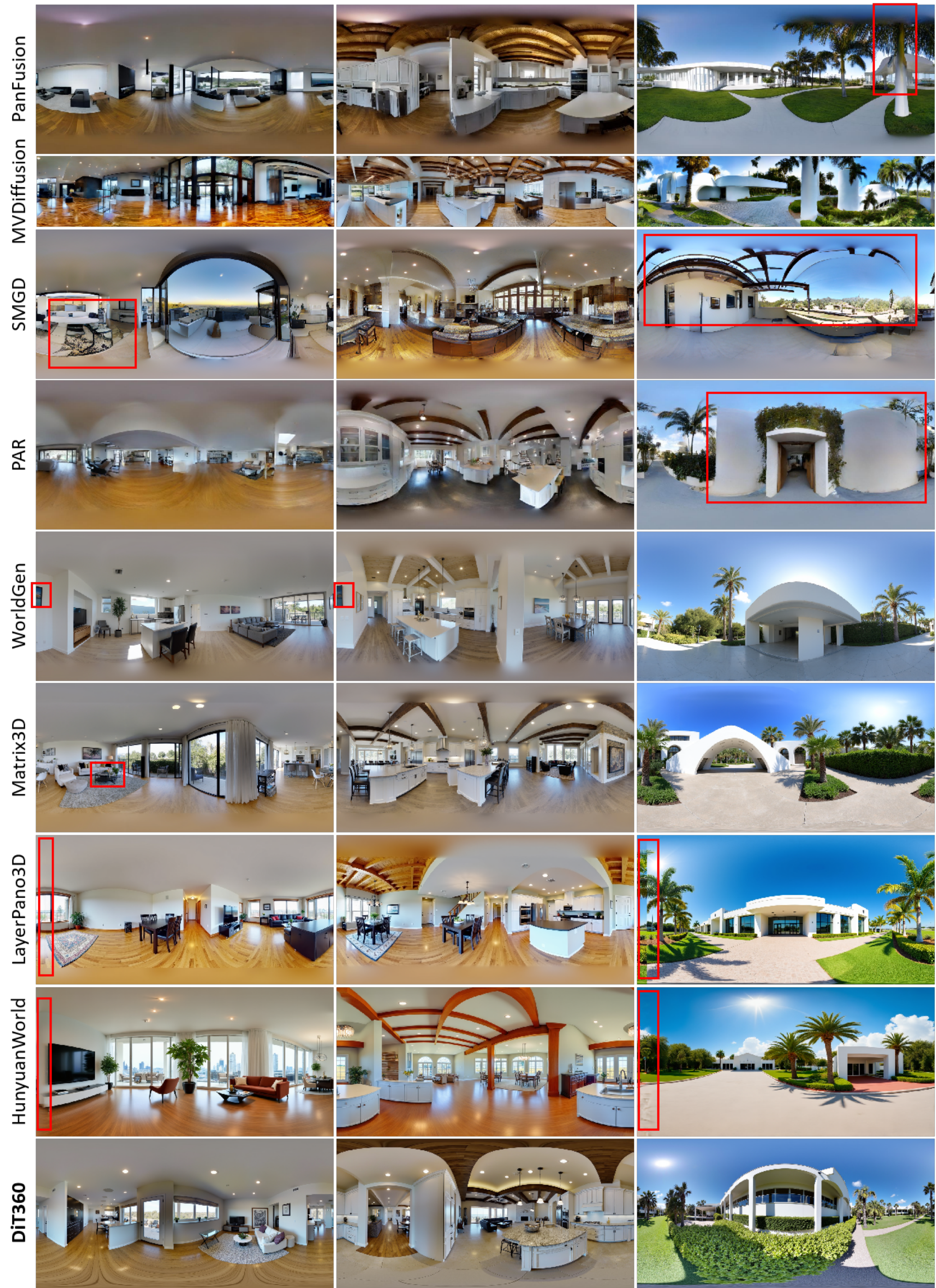}
    \caption{
    The full qualitative comparison on panorama generation. We highlight representative artifacts with red boxes.
    }
    \label{fig:app_comparison}
\end{figure}

In this section, we present the complete qualitative comparison of text-to-panorama generation results. 
As shown in \cref{fig:app_comparison}, our method demonstrates superior perceptual realism, producing sharper and more visually authentic panoramas. 
In addition, it achieves higher geometric fidelity by effectively handling distortions and preserving boundary continuity, whereas baseline methods often suffer from visible artifacts and structural inconsistencies.

\section{User Study}
\label{appendix:user_study}
\begin{table}[htbp]
\centering
\caption{User study results on text-to-panorama generation.}
\label{tab:user_study}
\resizebox{\textwidth}{!}{
\begin{tabular}{lcccc} 
\toprule
Methods & 
Text Alignment$\uparrow$ & 
Boundary Continuity$\uparrow$ & 
Realism$\uparrow$ & 
Overall Quality$\uparrow$ \\

\midrule
PanFusion     & 21.7\% & 19.6\% & 2.1\% & 0.3\% \\
Matrix-3D     & 24.1\% & 27.5\% & 23.7\% & 5.1\% \\
HunyuanWorld  & 25.9\% & 18.9\% & 10.4\% & 13.7\% \\
Ours          & \textbf{28.3\%} & \textbf{34.0\%} & \textbf{63.8\%} & \textbf{80.9\%} \\
\bottomrule
\end{tabular}
}
\end{table}

To further evaluate human preference, we conducted a user study comparing our method with several representative baselines~\citep{panfusion,matrix3d,hunyuanworld}.  
The study focused on four key aspects: text alignment, boundary continuity, realism, and overall quality.  
A total of 63 participants were asked to choose their preferred outputs from different methods on the test set.  
As shown in \cref{tab:user_study}, our method received the highest preference across all metrics, clearly demonstrating its superior ability to generate realistic panoramic images with faithful alignment and coherent boundaries.  

\section{More Results} 
\label{appendix:more_results}
\begin{figure}[t]
    \centering
    \includegraphics[width=\linewidth]{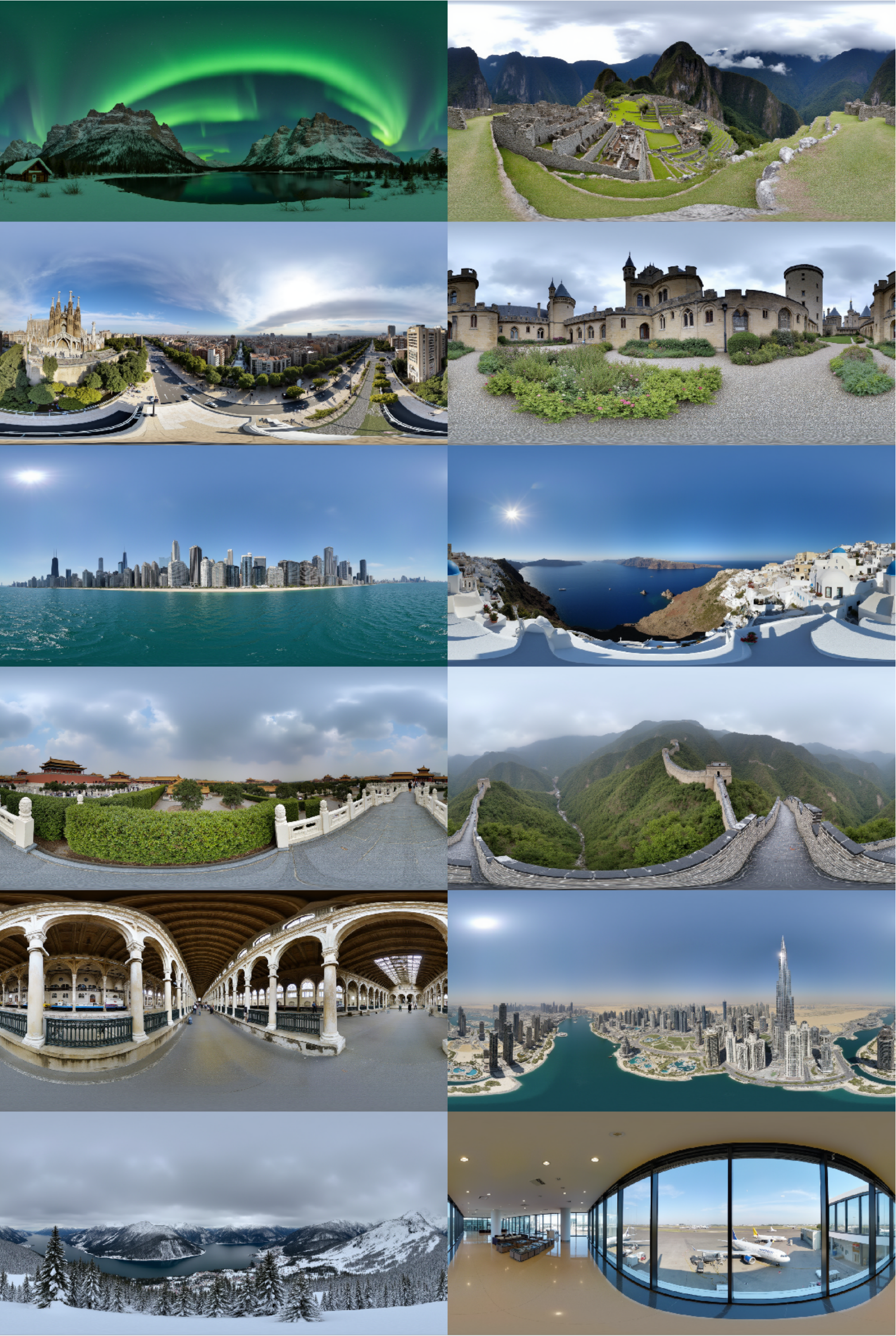}
    \caption{More results on text-to-panorama generation.} 
    \label{fig:more_results_1}
\end{figure}

\begin{figure}[t]
    \centering
    \includegraphics[width=\linewidth]{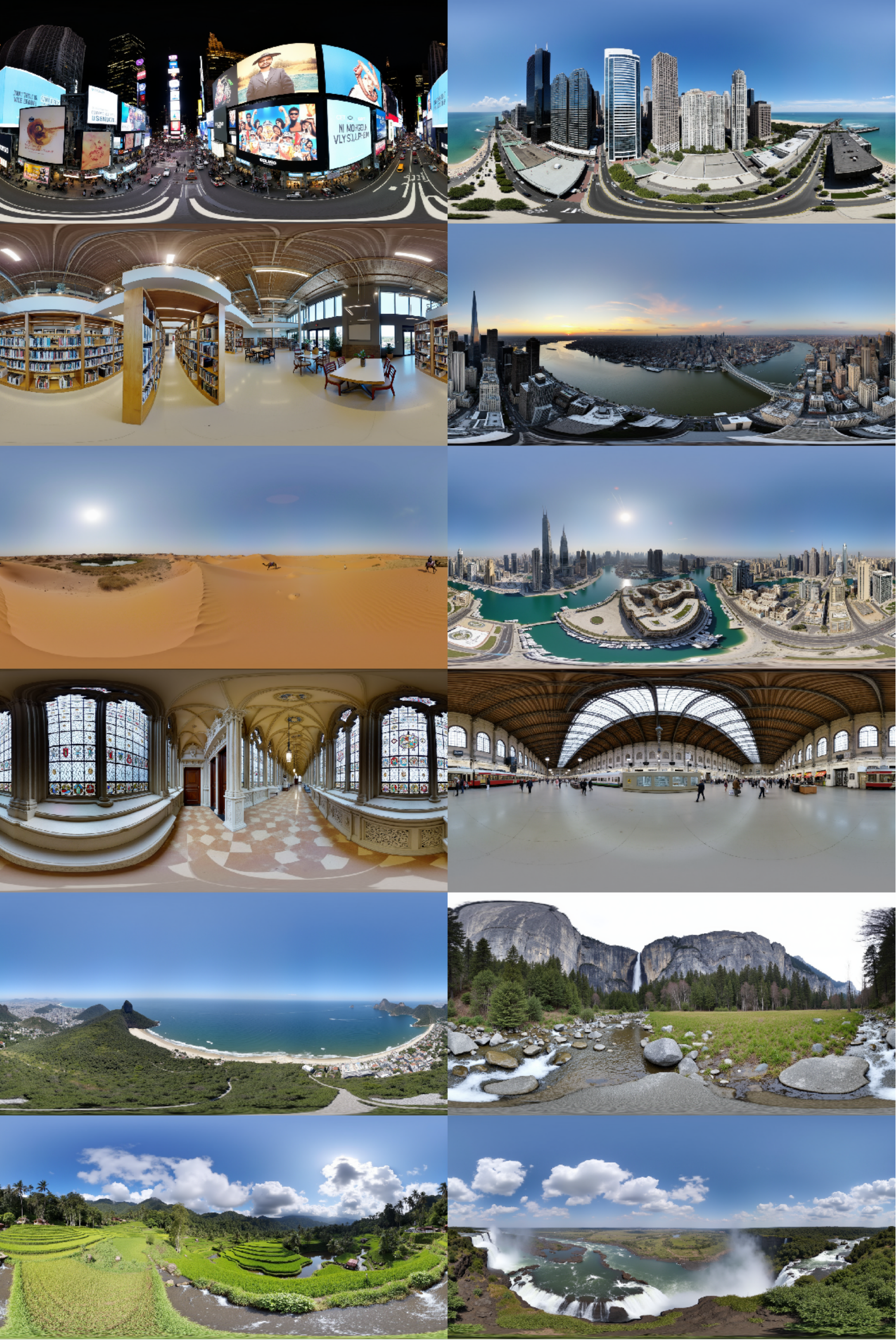}
    \caption{More results on text-to-panorama generation.} 
    \label{fig:more_results_2}
\end{figure}

We present additional results in~\cref{fig:more_results_1,fig:more_results_2} to further illustrate the performance of \textit{DiT360} on panoramic image generation. 
These examples demonstrate that the model consistently produces high-quality, semantically coherent, and visually detailed completions across a variety of scenes.

\section{Use of Large Language Models}
Large Language Models were used for minor grammar and style corrections only. 
All technical content, experiments, and conclusions were authored by the paper’s authors.

\section{Limitations and Future Work}

Despite the strong performance of \textit{DiT360} on panoramic image generation tasks, several limitations remain. 
The model’s effectiveness is constrained by the diversity and scale of available datasets, leading to suboptimal results in certain scenarios, such as those containing high-resolution human faces or intricate scene details. 
Future work will focus on collecting larger and more diverse high-quality datasets to further enhance the model’s generative capabilities and image resolution. 
Additionally, leveraging synthetic data to augment training samples can facilitate further advances in panoramic image generation. 
In the long term, extending the framework to three-dimensional scene generation and understanding represents a promising research direction.

% \section{Appendix}
% You may include other additional sections here.

\end{document}